\documentclass[10pt,twocolumn,letterpaper]{article}

\usepackage[pagenumbers]{cvpr} 

\usepackage{graphicx}
\usepackage{amsmath}
\usepackage{amssymb}
\usepackage{booktabs}
\usepackage[accsupp]{axessibility} 

\usepackage[pagebackref,breaklinks,colorlinks]{hyperref}

\usepackage[capitalize]{cleveref}
\crefname{section}{Sec.}{Secs.}
\Crefname{section}{Section}{Sections}
\Crefname{table}{Table}{Tables}
\crefname{table}{Tab.}{Tabs.}

\usepackage{lipsum}

\newcommand\blfootnote[1]{%
  \begingroup
  \renewcommand\thefootnote{}\footnote{#1}%
  \addtocounter{footnote}{-1}%
  \endgroup
}

\def\cvprPaperID{36} 
\def\confName{CVPR}
\def\confYear{2022}

\begin{document}

\title{Generalizing Adversarial Explanations with Grad-CAM}

\author{Tanmay Chakraborty, Utkarsh Trehan, Khawla Mallat\textsuperscript{\textsection}, and Jean-Luc Dugelay\\
EURECOM\\
Campus SophiaTech, 450 Route des Chappes, 06410, Biot, France.\\
{\tt\small \{chakrabo, trehan, mallat, dugelay\}@eurecom.fr}
}
\maketitle

\begingroup\renewcommand\thefootnote{\textsection}
\footnotetext{Khawla is now with SAP Security Research Labs France.}
\begin{abstract}
  Gradient-weighted Class Activation Mapping (Grad- CAM), is an example-based explanation method that provides a gradient activation heat map as an explanation for Convolution Neural Network (CNN) models. The drawback of this method is that it cannot be used to generalize CNN behaviour. In this paper, we present a novel method that extends Grad-CAM from example-based explanations to a method for explaining global model behaviour. This is achieved by introducing two new metrics, (i) Mean Observed Dissimilarity (MOD) and (ii) Variation in Dissimilarity (VID), for model generalization. These metrics are computed by comparing a Normalized Inverted Structural Similarity Index (NISSIM) metric of the Grad-CAM generated heatmap for samples from the original test set and samples from the adversarial test set. For our experiment, we study adversarial attacks on deep models such as VGG16, ResNet50, and ResNet101, and wide models such as InceptionNetv3 and XceptionNet using Fast Gradient Sign Method (FGSM). We then compute the metrics MOD and VID for the automatic face recognition (AFR) use case with the VGGFace2 dataset. We observe a consistent shift in the region highlighted in the Grad-CAM heatmap, reflecting its participation to the decision making, across all models under adversarial attacks. The proposed method can be used to understand adversarial attacks and explain the behaviour of black box CNN models for image analysis.
\end{abstract}
\vspace{-0.6cm}
\section{Introduction}
\label{sec:intro}

Automatic Face Recognition (AFR) systems are extremely useful tools in today's world, used in banking, automated border control, healthcare, biometrics, and security applications. Due to the high success rate of systems based on Deep Learning (DL), they are often preferred for these tasks. Computer Vision has found its application in numerous verticals since the success of convolutional neural networks \cite{NIPS2012_c399862d}. However with the increasing use of DL models in critical applications, it has been found that DL models are subject to attacks. These attacks in the form of some calculated mathematical noise on the input image cause the model to misclassify images \cite{szegedy2014intriguing}. These adversarial attacks display specific properties, i) They are not perceptible to the human eye, ii) They are controllable, and iii) Transferability, i.e., an attack designed for one model is capable of attacking multiple models \cite{demontis2019adversarial}.  

There is a number of literature on how to attack DL models successfully \cite{alparslan2020adversarial}. There are mainly two kinds of attacks: targeted and non-targeted attacks. Targeted attack makes a model predict a certain label for the adversarial example, while for non-targeted attacks the labels for adversarial examples are not important, as long as the model is wrong \cite{vakhshiteh2020adversarial}. These attacks can also be subdivided into black-box attacks and white-box attacks. Black-box attacks have no information about the target model, training procedure, architecture, whereas white-box attacks know the target model, training procedure, architecture, parameters. Research has also shown adversarial attacks can also be performed across spectrum, i.e., models can be fooled with examples whose spectrum are transposed ~\cite{bisogni2021adversarial}.  

As DL models become more vulnerable to adversarial attacks, it is desirable to have defensive measure and to look into models that have adversarial robustness. In ~\cite{Xie_2019_CVPR}, feature denoising has been proposed as a method to achieve models that are robust to a number of gradient based adversarial attacks. Adversarial attacks are often successful due to certain neurons which are more sensitive to changes than other neurons, a robust model can be achieved by identifying and controlling the sensitivity of these neurons ~\cite{9286885}. 

It is a common practice in DL to tune hyperparameters to improve model performance. The CKA-similarity algorithm was used to compare the hidden representations of broad and deep models~\cite{nguyen2020wide}. They found that when the model capacity is large compared to the training set, a block structure emerges, which shows that the models propagate the main component of their hidden representation. A study on effect of network width on adversarial robustness suggests that for similar parameters, wider network achieves better utility but worst adversarial stability \cite{NEURIPS2021_3937230d}. 

One of the classical ways to detect adversarial attack is by expanding the neural network architecture to have a binary sub-network that can classify regular examples from adversarial examples \cite{metzen2017detecting}. Statistical-based methods are also quite commonly found in the literature \cite{roth2019odds} \cite{grosse2017statistical} \cite{gao2021maximum}. More recent methods leverage explainability of machine learning and use SHAP based signatures to detect adversarial attacks \cite{fidel2020explainability}. Gradient based similarity computation is also proposed under explainable methods for adversarial example detection \cite{dhaliwal2018gradient}.  

Explainable AI concepts have been used to analyze adversarial attacks, and their defense strategies~\cite{klawikowska2020explainable}. They concluded that deep networks are still vulnerable to adversarial attacks and are often not well prepared for intentional input disruption. Visual analysis based adversarial explainability is also found in the related literature \cite{cantareira2021explainable}. They show that a model precepts adversarial examples differently. Another visual analysis method to explain why models fail observes the flow path of an adversarial example compared to a regular example and produces heatmap, and feature level map~\cite{8802509}. Overall visual analysis of adversarial attacks and why networks actually fail is still under explored and not generalized.

In this work, we attempt to answer three fundamental questions. \vspace{-0.2cm}
\begin{itemize} \item Why do neural networks (wide vs. deep) fail under adversarial settings? \vspace{-0.2cm}\item Are certain architectures better than others? \vspace{-0.2cm}\item Can we find metrics that can help explain the behaviour (functioning) of CNN models when under adversarial attacks?
\end{itemize}\vspace{-0.2cm}

We leverage advances in explainability to take a look at what the model sees layer by layer using Gradient-Weighted Class Activation Mapping (Grad- CAM). It is an example-based explanation method, and it does not provide insights into global model behaviour. The other explanation methods for CNNs, such as KernelSHAP~\cite{lundberg2017unified}, also do not provide generalized results for CNNs. 

To address some of the above issues, we should understand the global model behaviour. Therefore, we introduce two new metrics, (i) Mean Observed Dissimilarity (MOD) and (ii) Variation in Dissimilarity (VID), for model generalization using Grad- CAM individual heatmap comparisons. We use the Normalized Inverted Structural Similarity Index (NISSIM) metric for the Grad- CAM generated heatmap of the samples from the original test set and the samples from the adversarial test set. Thus, we extend an example-based method into a global explanatory method.

For our case study, we take five out-of-the-box models, of which three are deep models (VGG16, ResNet50, ResNet101) and two wide models (InceptionNetV3, XceptionNet). We train all these models with 50 classes selected randomly from the face recognition dataset VGGFace2. We also attack these models using the Fast Gradient Sign Method (FGSM) for different noise~\cite{goodfellow2014explaining}. We generate results for both black-box attacks (attacks generated using ResNet50 and applied to all models) and a white-box attack (attacks generated for each model). For further analysis, we visualize these models layer by layer output using Grad-CAM for both with and without adversarial attacks. 

As a result, we observed a global pattern displayed by all models. The shifting in the region of participation can be defined as when a model sees adversarial examples. Some parts of the input image no longer participate in the decision-making, while new parts do participate. Models are not robust to these changes. These changes are not deterministic, and given an adversarial example, there is no way to tell how it will affect the shift. This dissimilarity is why models fail when attacked, and the extent of this dissimilarity can be quantified with MOD and VID metrics. These metrics are also desirable for robust training, where the goal is to protect a model from adversarial attacks. To be noted that robust training is out of the scope of this work. The goal of this work is global explainability for CNN under an adversarial setting. 

To summarize our contribution, we extend Grad-CAM, an example-based explanation method, to a global model explanation method by introducing dissimilarity metrics computed with NISSIM. We define the shifting in the region of participation for CNNs when they are under adversarial attacks. Moreover, we generalize this behaviour through our experiments across different models. 
\vspace{-0.2cm}
\section{Background}
We will use five established CNN models with transfer learning pre-trained on imageNet dataset for our purpose. The adversarial examples will be generated using FGSM for both black-box and white-box attacks.

\subsection{Neural networks}

\paragraph{VGG16} VGG16 is a deep convolutional neural network with 16 layers. The authors studied the effect of deeper layers while keeping the convolution kernels of size 3x3~\cite{simonyan2014very}. 
\vspace{-0.8cm}
\paragraph{ResNet50 \& ResNet101}
ResNets are versatile architectures that introduce residual blocks, skip connections, and shortcuts. They are considered very deep neural networks~\cite{he2015deep}. 
\vspace{-0.4cm}
\paragraph{InceptionNet}
InceptionNet is a wide convolutional neural network with factorized convolutions and aggressive regularization. The network is wider essentially than deeper. Also, with deeper networks, there were vanishing gradients. To prevent that the authors introduced auxiliary classifiers in between. They essentially applied softmax to the outputs of two inception modules and computed an auxiliary loss over the same labels. The total loss function is a weighted sum of the auxiliary loss and the real loss~\cite{szegedy2016rethinking}. 
\vspace{-0.4cm}
\paragraph{XceptionNet}
XceptionNet is another wide network that has been developed from InceptionNet by applying depth-wise separable convolutions. The number of parameters in XceptionNet is similar to InceptionNet, but the performance gain is due to more efficient use of model parameters~\cite{chollet2017xception}.
\vspace{-0.1cm}
\subsection{Fast Gradient Sign Method}
Creating adversarial examples requires us to add noise to the input image. FGSM is a method of generating noise in the direction of the cost function gradient concerning the data~\cite{goodfellow2014explaining}. The noise is then controlled by a parameter $\epsilon$. Given original input image $x$, label $y$, model parameter $\theta$, and loss $J$. We can write $adv_x = x+ \epsilon * sign(\nabla_x J(\theta,x,y))$ this gives us the perturbations. 




\subsection{Grad-CAM}
Grad-CAM is an example based model agnostic explanation tool for CNN that uses the gradient information of the target object and how it flows through a network to create coarse localization heatmaps for visual analysis~\cite{selvaraju2017gradcam}. The heatmap produced by Grad-CAM tells clearly for an image, which parts are under focus and considered by the CNN to come to a decision. Blue parts of the heatmap indicate no participation and red parts indicate high participation. 
\vspace{-0.2cm}
\section{Metrics}
In this section, we define the different metrics we introduce namely NISSIM, MOD, and VID.

\subsection{Normalized Inverted Structural Similarity Index}
Normalized Inverted Structural Similarity Index (NISSIM) metric is calculated from Structural Similarity Index (SSIM) by inverting the range and then normalizing it. SSIM is a metric that focuses on similarity, thus inverting it gives us dissimilarity. SSIM bounds to (-1, 1], where -1 means dissimilar while 1 means similar, and NISSIM bounds to (0, 1] where 0 means similar and 1 means dissimilar. Ideally we want this value as close to 0 as possible. 
\vspace{-0.3cm}
\begin{align}
 NISSIM_i = \frac{1 - SSIM_i}{2}
\end{align}
\vspace{-0.7cm}
\subsection{Mean Observed Dissimilarity}
Mean Observed Dissimilarity (MOD), is the mean of the NISSIM dissimilarity over the adversarial test set for similar levels of attack. So for every adversarial set $\mathit{X^*}$, calculate NISSIM value for all samples in that set, and divide by the total number of samples. This metric is bounded in between (0,1], such that 0 indicates total similarity while 1 indicates total dissimilarity. This metric is an indication on how much the model is robust under adversarial settings but still inconclusive. As models may be stable to shift, thus the need for us to look into adversarial stability with variation in dissimilarity metric.  
\vspace{-0.3cm}
\begin{align}
 MOD_{advset} = \frac{1}{N}\sum{NISSIM_i}
\end{align}

\subsection{Variation in Dissimilarity}
Variation in Dissimilarity (VID) is the variance of the NISSIM metric over the adversarial set over different levels of attack $\mathit{eps}$ for a model, this shows the distribution of the attack on the model when different levels of attack are performed. This metric indicates the stability of a model under adversarial settings. Ideally, we would want the distribution to be stable for different levels of attack.

\vspace{-0.4cm}
\begin{align}
 m_{h} = \frac{1}{eps}\sum{NISSIM_{eps}}\\
 VID = \sqrt{\frac{\sum({NISSIM_{eps} - m_{h}})^2}{eps}}
\end{align}
\vspace{-0.7cm}

\section{Case Study}

We are conducting a case study on AFR systems. AFR systems use DL for many sensitive tasks such as biometrics and security monitoring. It is common for attackers to attack these sensitive systems. Since CNN models can be outwitted with adversarial examples, it is desirable to find explanations of how these models work and why they fail. 

We choose three deep models, namely VGG16, ResNet50, and ResNet101, and two wide models, InceptionNetV3 and XceptionNet, to recognize faces from input images.

\subsection{Case study pipeline}
For our case study, we will use a subset of 50 randomly chosen classes from VGGFace2 dataset. VGGFace2 is a large-scale face image dataset created by scraping images from Google image search. The dataset houses 3.31 million images of 9131 classes~\cite{cao2018vggface2}. 

First, we preprocess the dataset to align and crop the faces. Then, the dataset is split into 80\% training, 10\% testing, and 10\% validation sets. The fully connected layers are trained for 60 epochs on the training set and the best weight according to the validation accuracy is stored. 
Once the training step is completed, the stored models are loaded and used to generate perturbations from the test set using FGSM. Then the test set is attacked with different values of $\epsilon$ from the stored perturbations and these counterexamples are stored as perturbed test sets (one test set is created for each epsilon). Then the five models are tested with these sets and the performance is noted. 
First, FGSM attacks were generated using the ResNet50 model and the attacks were transferred to the other models. This verified the transferability of the attacks. Then, attacks were also generated for each model and the respective models were also attacked. 
Finally, Grad- CAM was used to generate heatmaps for every layer in each model and each $\epsilon$ in the perturbed test set. Subsequently, these heatmaps were analysed by computing the metrics NISSIM, MOD, and VID. We observed a consistent shifting behaviour of the region highlighted by Grad-CAM among all models that were attacked. This allowed us to generalise the behaviour of CNNs under adversarial environment.

\subsection{Performance degradation}
Two types of performance analysis have been performed, one without preprocessing the data and the other with preprocessing. We calculated the Peak Signal to Noise Ratio (PSNR) metric shown in Fig.~\ref{fig:psnr}, with the goal to To understand how much noise we introduce during the attacks. A successful attack is given in Fig.~\ref{fig:examplefull}, to illustrate the working of adversarial attack with FGSM. The degradation of the overall accuracy of each model after an attack for different values of $\epsilon$ is given in table~\ref{tab:fullscale} for the set without preprocessing and table~\ref{tab:fullscalepre} for the set with preprocessing. These two tables indicate that the FGSM attacks on the model was successful, and we were able to fool the deep models. These attacks are black-box attacks, meaning that the attacks were generated using ResNet50 and all models were attacked. We can observe a pattern that wide models fail more than deep models as the $\epsilon$ increases. We also observed that preprocessing the input gave some performance benefits for all models, but the failure trend remained the same. 

Finally, we take a look at the white-box attack for a particular example and see how the models perform. The results are given in the Fig.~\ref{fig:vgg16_grad}, ~\ref{fig:resnet50_grad}, ~\ref{fig:resnet101_grad}, ~\ref{fig:inceptionnet_grad}, and ~\ref{fig:xception_grad}. The original image given for testing is of ABD and where the model was wrong, it identified ABD as Buckley. Again, we portray that the FGSM attack was successful and the models failed. 


         
         
         
         

\begin{table}[htpb]
    \centering
    \caption{Overall accuracy obtained with full-scale attack for data without pre-processing, depicting the success of FGSM attack}
    \scalebox{0.6}{
    \begin{tabular}{||c | c | c | c | c | c ||}
    \hline
         Model & $\epsilon = 0.000$ & $\epsilon = 0.010$ & $\epsilon = 0.050$ & $\epsilon = 0.075$ & $\epsilon = 0.100$ \\
         \hline
         VGG16 & 91.89\% & 91.55\% & 90.87\% & 89.86\% & 89.86\% \\
         ResNet50 & 90.54\% & 91.89\% & 88.51\% & 78.04\% & 70.60\% \\
         ResNet101 & 90.87\% & 92.22\% & 89.18\% & 75.33\% & 61.82\% \\
         InceptionNet v3 & 88.51\% & 88.51\% & 57.09\% & 35.13\% & 25.33\%\\
         XceptionNet & 92.56\% & 92.56\% & 81.41\% & 69.59\% & 56.41\%\\
         \hline
    \end{tabular}}
    \vspace{-0.3cm}
    \label{tab:fullscale}
\end{table}
\vspace{-0.2cm}
\begin{table}[htpb]
    \centering
    \caption{Overall accuracy obtained with full-scale attack for data with pre-processing, depicting the success of FGSM attack}
    \scalebox{0.6}{
    \begin{tabular}{||c | c | c | c | c | c ||}
    \hline
         Model & $\epsilon = 0.000$ & $\epsilon = 0.010$ & $\epsilon = 0.050$ & $\epsilon = 0.075$ & $\epsilon = 0.100$ \\
         \hline
         VGG16 & 87.54\% & 85.60\% & 85.21\% & 84.82\% & 84.43\% \\
         ResNet50 & 94.16\% & 94.55\% & 90.66\% & 85.21\% & 75.09\% \\
         ResNet101 & 93.38\% & 93.38\% & 87.15\% & 84.82\% & 77.43\% \\
         InceptionNet v3 & 85.21\% & 84.82\% & 66.92\% & 47.85\% & 29.96\%\\
         XceptionNet & 89.49\% & 90.27\% & 71.20\% & 59.14\% & 45.13\%\\
         \hline
    \end{tabular}}
    \label{tab:fullscalepre}
\end{table}

\begin{figure}[htpb]
    \centering
    \includegraphics[scale=0.35]{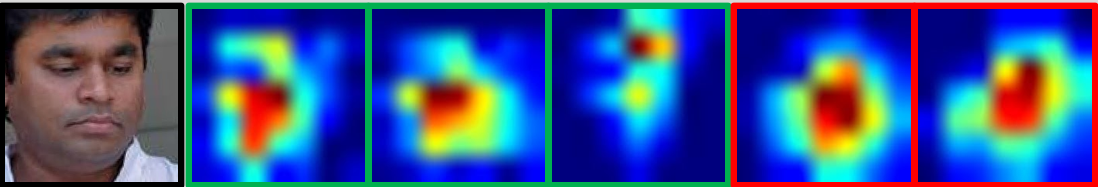}
    \caption{The observed shifting in the region of participation highlighted by Grad-CAM}
    \label{fig:shifting}
\end{figure}

\begin{figure}[htpb]
    \centering
    \vspace{-0.3cm}
    \includegraphics[scale=0.28]{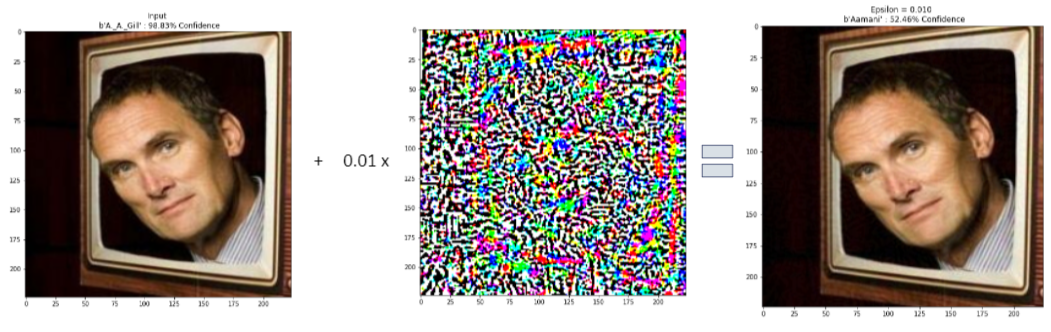}
    \caption{Example of a successful full-scale attack with $\epsilon = 0.01$ }
    \label{fig:examplefull}
\end{figure}
\begin{figure}[h]
    \centering
    \includegraphics[scale=0.6]{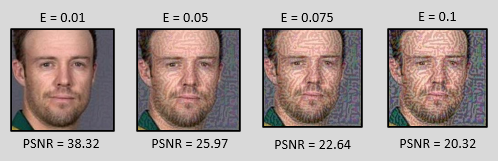}
    \caption{PSNR table for noise quantification.}
    \label{fig:psnr}
\end{figure}
\vspace{-0.5cm}

\begin{figure*}[htbp]
    \centering
    \subfloat[\centering $\epsilon = 0.01$]{{\includegraphics[scale=0.24]{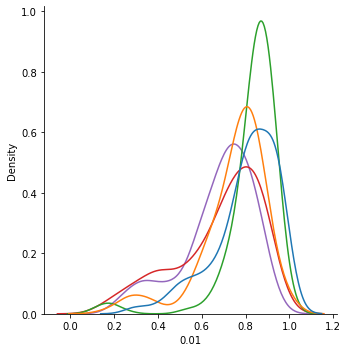} }}%
    \hfill
    \subfloat[\centering  $\epsilon = 0.05$]{{\includegraphics[scale=0.24]{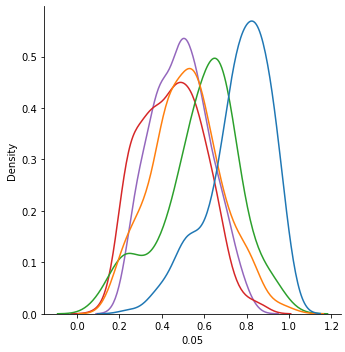} }}%
    \hfill
    \subfloat[\centering  $\epsilon = 0.075$]{{\includegraphics[scale=0.24]{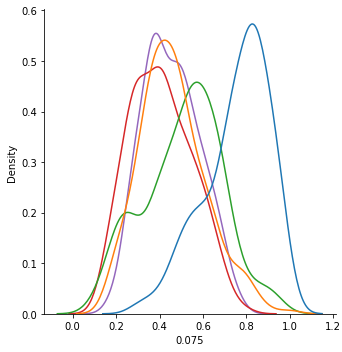} }}%
    \hfill
    \subfloat[\centering  $\epsilon = 0.1$]{{\includegraphics[scale=0.24]{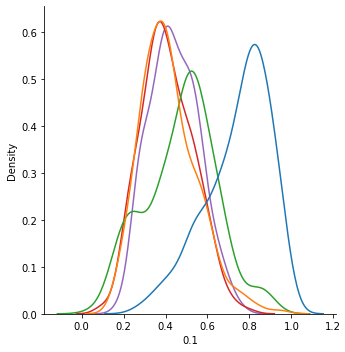} }}%
    \hfill
    \subfloat[\centering  legend]{{\includegraphics[scale=0.9]{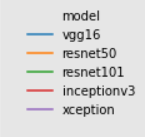} }}%
    \caption{VID distribution for different $\epsilon$, the distributions show the variation of shift in region of participation for each model.}%
    \label{fig:vid0}%
\end{figure*}

\subsection{Analysis and Discussions}
For this detailed analysis, we will examine two types of heatmaps generated using Grad-CAM. First, layer by layer heatmaps for all models and epsilon values using the white-box attack to better understand how each model behaves. And second, heatmaps for black-box attacks to understand if the behaviours are consistent and if they can be generalized.

\begin{figure}[h]
    \centering
    \includegraphics[scale=0.27]{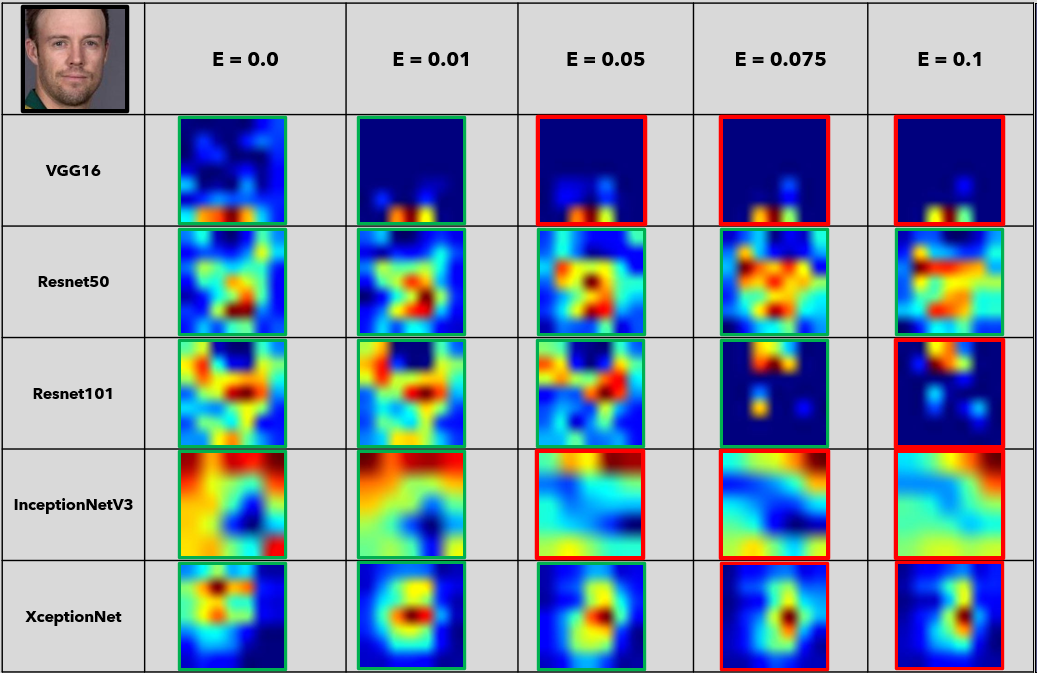}
    \caption{Black box attack example, depicting the shifting behaviour in the region of participating as $\epsilon$ increases}
    \label{fig:bb-example}
\end{figure}
\vspace{-0.6cm}
\subsubsection{Explaining white-box attacks}
\vspace{-0.2cm}
For generating the explanation heatmaps for every layer of a given model with Grad-CAM, we must first remove the last classification layer so that we can access the gradients. On the heatmap, blue means no involvement and red means the highest involvement in the decision-making. The plots are read from bottom right to top left, where bottom right is the first layer and top left is the last layer. The goal of these plots are to observe the shifting region of participation behaviour visually as the model sees them in each layer. 

VGG 16 given in Fig.~\ref{fig:vgg16_grad}, we can observe clearly that all the attacks were successful and illustrates a clear shift of participating regions as the $\epsilon$ increases. ResNet50 given in Fig.~\ref{fig:resnet50_grad}, the number of layers are too many to pin point out some example, yet if observed very carefully the hidden layers as the $\epsilon$ increases, we can find a shifting in the region of participation.
In ResNet101 given in Fig.~\ref{fig:resnet101_grad}, it seems more resilient there are some observable region shifts, but overall much less. InceptionNet v3 given in Fig.~\ref{fig:inceptionnet_grad}, seems to have learnt something different, the focus was more on forehead than face, but the overall shifting is much higher for this model, we even see focus regions getting inverted as the $\epsilon$ increases. For XceptionNet given in Fig.~\ref{fig:xception_grad}, the phenomenon is more clear, some regions get expanded, and background areas are being highlighted.

\begin{figure}[h]
    \centering
    \includegraphics[scale=0.27]{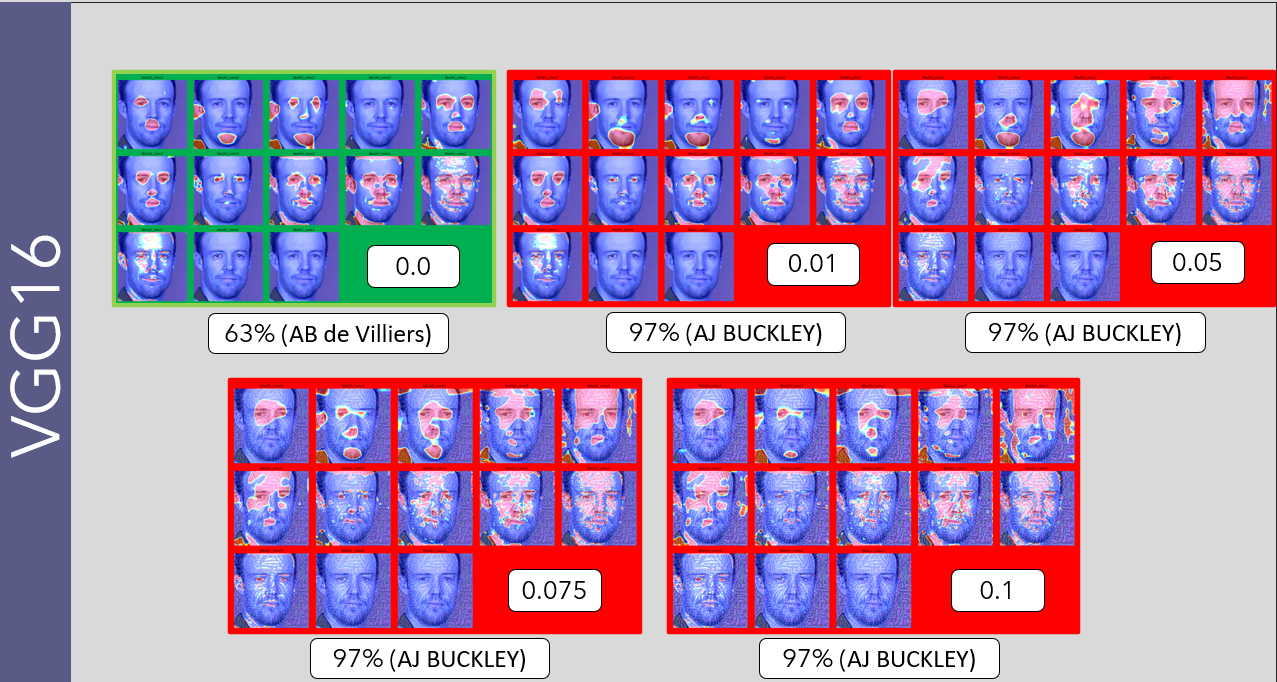}
    \caption{Layer by layer visualization of Grad-CAM heatmaps of VGG 16, depicting shifting region of participating as $\epsilon$ increases.}
    \label{fig:vgg16_grad}
\end{figure}

\begin{figure}[h]
    \centering
    \includegraphics[scale=0.28]{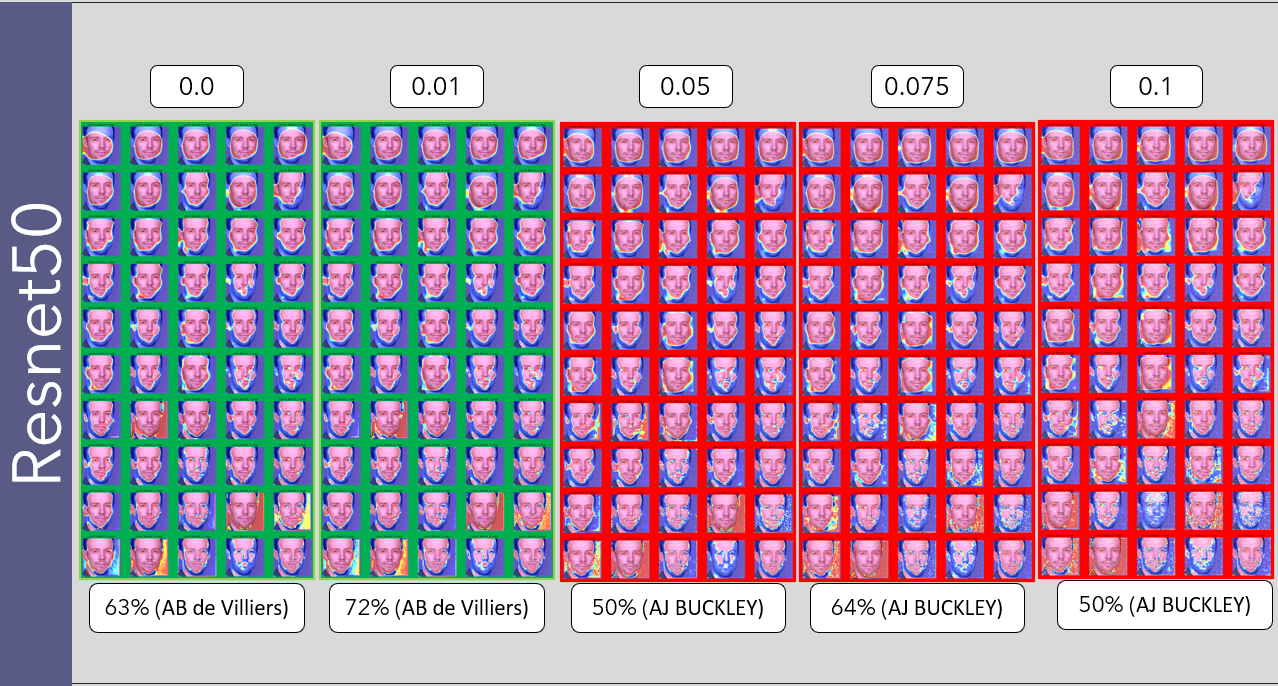}
    \caption{Layer by layer visualization of Grad-CAM heatmaps of ResNet50, depicting shifting in the region of participating as $\epsilon$ increases.}
    \label{fig:resnet50_grad}
\end{figure}

\vspace{-0.2cm}
\begin{figure}[h]
    \centering
    \includegraphics[scale=0.28]{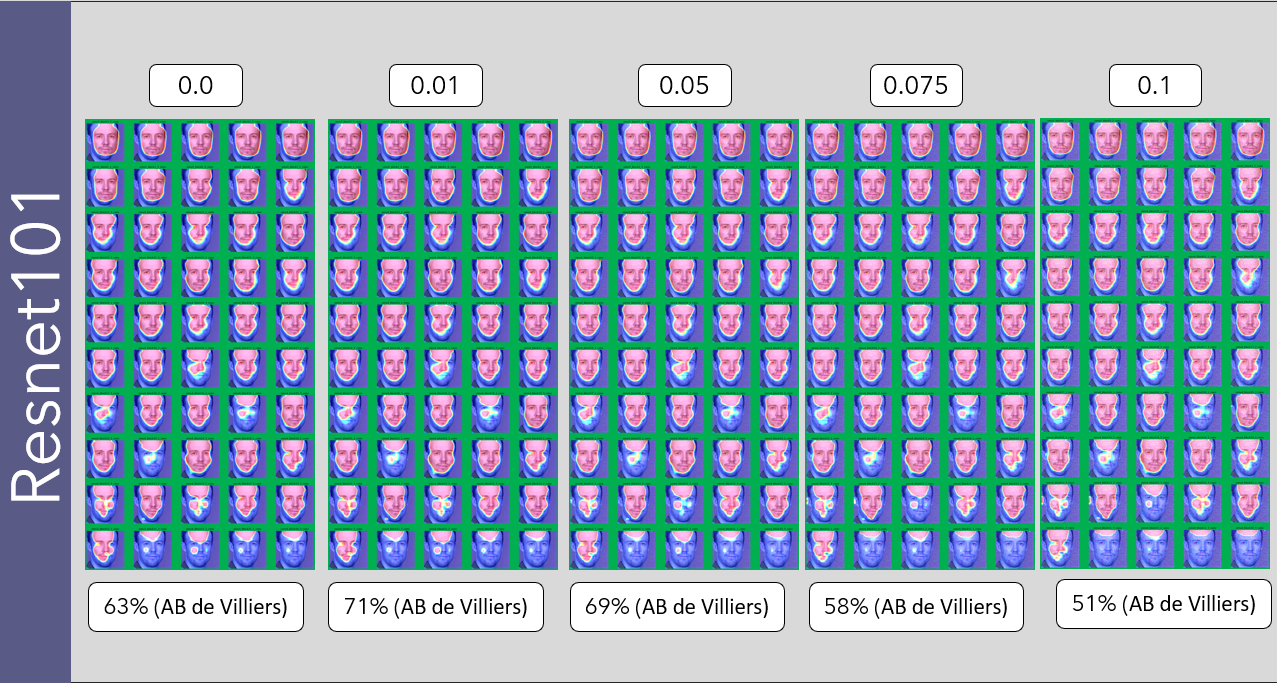}
    \caption{Layer by layer visualization of Grad-CAM heatmaps of ResNet101, depicting shifting in the region of participating as $\epsilon$ increases.}
    \label{fig:resnet101_grad}
\end{figure}

\begin{figure}[h]
    \centering
    \includegraphics[scale=0.28]{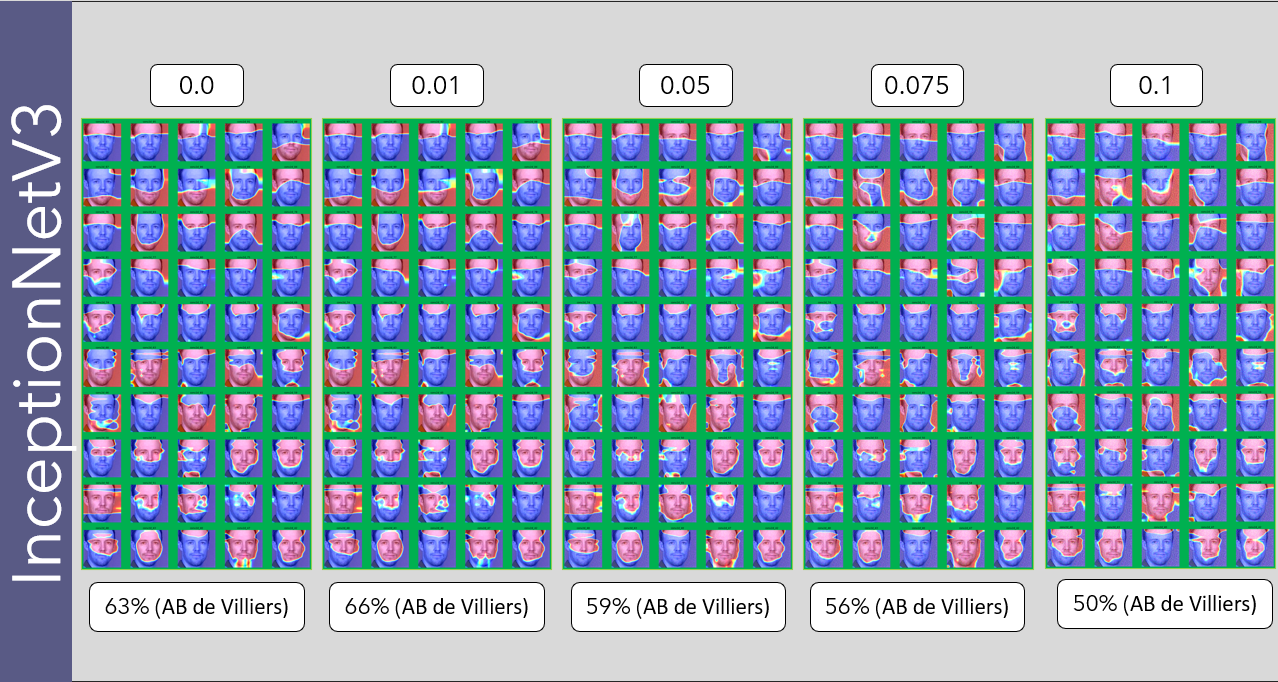}
    \caption{Layer by layer visualization of Grad-CAM heatmaps of InceptionNet v3, depicting shifting in the region of participating as $\epsilon$ increases.}
    \label{fig:inceptionnet_grad}
\end{figure}

\begin{figure}[h]
    \centering
    \includegraphics[scale=0.27]{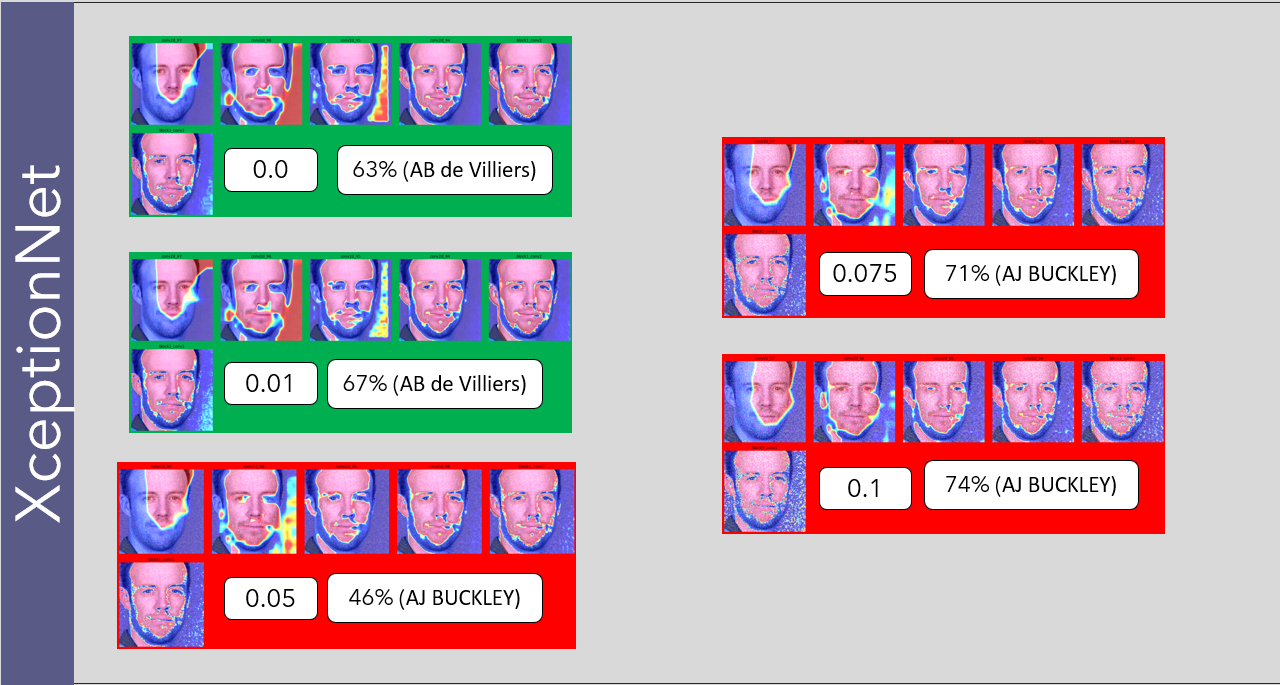}
    \caption{Layer by layer visualization of Grad-CAM heatmaps of XceptionNet, depicting shifting in the region of participating as $\epsilon$ increases.}
    \label{fig:xception_grad}
\end{figure}

\subsubsection{Explaining black-box attacks}
For the back-box attack, we extract the raw explanation heatmap with the goal of observing more clearly the shifting region of participation phenomenon as given in Fig.~\ref{fig:bb-example}. For each model we can observe shifting as $\epsilon$ increases. Which is why we need to quantify this shifting to generalize the phenomenon across models. 
\vspace{-0.1cm}
\subsection{MOD and VID computation}
We take the raw heatmap as computed in the black-box attack for all adversarial groups and compute the NISSIM metric. This metric requires two heatmaps to compute structural dissimilarity. We use the heatmap obtained from the original image (without adversarial perturbation) as our ground truth heatmap, i.e., what the model expects to see in order to make a decision, then the second heatmap is generated from the adversarially attacked image, and we create a dataframe with all the NISSIM values for the entire test set comparing with the adversarial test set for all values of $\epsilon$. 

Next, we compute the MOD metric from this set for all models given in the table~\ref{tab:mod}. The principal idea behind this table is to quantify the observed shifting for every level of $\epsilon$ value. We can see that the first column is the ground truth, all heatmaps are similar, so the MOD value is 0. The main observation to keep in mind is, as $\epsilon$ increases, the dissimilarity increases, indicating that the focus of the model is diverted when it is presented an adversarial example, this value indicates that the more the examples differ, the more likely the model will fail. We can also use this metric to explain the performance of VGG16. Since the shift was smaller, the model was less likely to fail. We also find that deep networks perform better than wide networks for similar shifts.
\vspace{-0.1cm}
\begin{table}[htpb]
    \centering
    \caption{MOD values obtained for different $\epsilon$ values, quantifying the observed shifting for every level of $\epsilon$}
    \scalebox{0.7}{
    \begin{tabular}{||c | c | c | c | c | c ||}
    \hline
         Model & $\epsilon = 0.000$ & $\epsilon = 0.010$ & $\epsilon = 0.050$ & $\epsilon = 0.075$ & $\epsilon = 0.100$ \\
         \hline
         VGG16 & 0 & 0.098875 & 0.117416 & 0.121993 & 0.127394 \\
         ResNet50 & 0 & 0.127743 & 0.241743 & 0.269364 & 0.292002 \\
         ResNet101 & 0 & 0.084589 & 0.208446 & 0.245334 & 0.262114 \\
         InceptionNet v3 & 0 & 0.160696 & 0.275883 & 0.296766 & 0.295898\\
         XceptionNet & 0 & 0.162345 & 0.255316 & 0.272804 & 0.27981\\
         \hline
         \hline
         Mean shift & 0 & 0.12685 & 0.219761 & 0.241252 & 0.251444\\
         \hline
    \end{tabular}}
    \label{tab:mod}
\end{table}
\vspace{-0.2cm}
Next, we observe the distribution of attacks with VID to understand the stability of each model across $\epsilon$ values. This distribution shows the variation of shift in the region of participation to the decision-making of a model. The goal of the distributions given in Fig.~\ref{fig:vid0} is to observe the stability of each model under adversarial settings. The main idea, that is examined here, is that the lower the shift in distribution, more the model is robust to adversarial attacks. The shifting in distribution reflects distribution of the shift in the region of participation for different models. We see that the distribution for VGG16 does not shift much over the course of the attack, while the distribution for the other models is scattered and shifts to the left. This indicates that VGG16 is a stable model for this task, over the other models. The two metrics together provide valuable insights into the performance and stability of CNNs under adversarial attacks. Therefore, it is desirable to compute these metrics while we perform defence strategies such as robust training.
\vspace{-0.5cm}
\subsection{Shifting behaviour in the region of participation}

From our extensive result analysis, we can generalize the shifting in the region of participation as seen in Fig.~\ref{fig:shifting}. When adversarial examples are seen by a model, there are some parts of the input for which the participation is discarded in decision making, and some other parts start playing a role in the decision process yielding a wrong prediction. We see a shift in the focus of the model in different directions, sometimes backgrounds get highlighted, other times, participation region expands or shrinks. This phenomenon is still non-deterministic in nature, i.e. given an adversarial example it is not possible to predict how focus will be shifted. We also observe that wide networks are more susceptible to this kind of shifting under adversarial attacks. Which more concretely explains our performance degradation trend. Deeper models are much robust to this changes, for similar amount of shift, deeper models provide better performance than wider models. 
\vspace{-0.2cm}

\section{Conclusions}
\vspace{-0.1cm}
Neural networks, the core component of a DL system, are essentially a black-box. Very little is known about why these models work, and even less about why they do not. This article attempts to understand popular deep and wide CNN architectures from the perspective of an adversarial attack. From our case study with AFR, we could conclude that neural networks fail because of a shifting behaviour in the region of participation to the decision-making, when the model sees adversarial examples, its focus changes and it now sees a different hidden representation. It appears that deep architectures are more robust to the shifting behaviour in the region of participation, even if they exhibit this behaviour, their degradation is much lower compared to wide networks. This is evident from the various metrics we computed. It is also observed that VGG16 is more stable in an adversarial environment than the other models we have studied. We introduce the NISSIM, MOD, and VID metrics to generalize adversarial behaviour of CNN by quantifying the shifting behaviour in the region of participation. Thus, we extend the example-based explanatory method to a global method for explaining model behaviour. 
\vspace{-0.1cm}
\blfootnote{\textbf{Acknowledgement} This work has been partially supported by the European CHIST-ERA program via the French National Research Agency (ANR) within the XAIface project (grant agreement CHIST-ERA-19-XAI-011).
}

{\small
\bibliographystyle{ieee_fullname}

\begin{thebibliography}{10}\itemsep=-1pt

\bibitem{alparslan2020adversarial}
Yigit Alparslan, Ken Alparslan, Jeremy Keim-Shenk, Shweta Khade, and Rachel
  Greenstadt.
\newblock Adversarial attacks on convolutional neural networks in facial
  recognition domain.
\newblock {\em arXiv preprint arXiv:2001.11137}, 2020.

\bibitem{bisogni2021adversarial}
Carmen Bisogni, Lucia Cascone, Jean-Luc Dugelay, and Chiara Pero.
\newblock Adversarial attacks through architectures and spectra in face
  recognition.
\newblock {\em Pattern Recognition Letters}, 147:55--62, 2021.

\bibitem{cantareira2021explainable}
Gabriel~D Cantareira, Rodrigo~F Mello, and Fernando~V Paulovich.
\newblock Explainable adversarial attacks in deep neural networks using
  activation profiles.
\newblock {\em arXiv preprint arXiv:2103.10229}, 2021.

\bibitem{cao2018vggface2}
Qiong Cao, Li Shen, Weidi Xie, Omkar~M Parkhi, and Andrew Zisserman.
\newblock Vggface2: A dataset for recognising faces across pose and age.
\newblock In {\em 2018 13th IEEE international conference on automatic face \&
  gesture recognition (FG 2018)}, pages 67--74. IEEE, 2018.

\bibitem{chollet2017xception}
François Chollet.
\newblock Xception: Deep learning with depthwise separable convolutions, 2017.

\bibitem{demontis2019adversarial}
Ambra Demontis, Marco Melis, Maura Pintor, Matthew Jagielski, Battista Biggio,
  Alina Oprea, Cristina Nita-Rotaru, and Fabio Roli.
\newblock Why do adversarial attacks transfer? explaining transferability of
  evasion and poisoning attacks.
\newblock In {\em 28th $\{$USENIX$\}$ Security Symposium ($\{$USENIX$\}$
  Security 19)}, pages 321--338, 2019.

\bibitem{dhaliwal2018gradient}
Jasjeet Dhaliwal and Saurabh Shintre.
\newblock Gradient similarity: An explainable approach to detect adversarial
  attacks against deep learning.
\newblock {\em arXiv preprint arXiv:1806.10707}, 2018.

\bibitem{fidel2020explainability}
Gil Fidel, Ron Bitton, and Asaf Shabtai.
\newblock When explainability meets adversarial learning: Detecting adversarial
  examples using shap signatures.
\newblock In {\em 2020 international joint conference on neural networks
  (IJCNN)}, pages 1--8. IEEE, 2020.

\bibitem{gao2021maximum}
Ruize Gao, Feng Liu, Jingfeng Zhang, Bo Han, Tongliang Liu, Gang Niu, and
  Masashi Sugiyama.
\newblock Maximum mean discrepancy test is aware of adversarial attacks.
\newblock In {\em International Conference on Machine Learning}, pages
  3564--3575. PMLR, 2021.

\bibitem{goodfellow2014explaining}
Ian~J Goodfellow, Jonathon Shlens, and Christian Szegedy.
\newblock Explaining and harnessing adversarial examples.
\newblock {\em arXiv preprint arXiv:1412.6572}, 2014.

\bibitem{grosse2017statistical}
Kathrin Grosse, Praveen Manoharan, Nicolas Papernot, Michael Backes, and
  Patrick McDaniel.
\newblock On the (statistical) detection of adversarial examples.
\newblock {\em arXiv preprint arXiv:1702.06280}, 2017.

\bibitem{he2015deep}
Kaiming He, Xiangyu Zhang, Shaoqing Ren, and Jian Sun.
\newblock Deep residual learning for image recognition, 2015.

\bibitem{klawikowska2020explainable}
Zuzanna Klawikowska, Agnieszka Miko{\l}ajczyk, and Micha{\l} Grochowski.
\newblock Explainable ai for inspecting adversarial attacks on deep neural
  networks.
\newblock In {\em International Conference on Artificial Intelligence and Soft
  Computing}, pages 134--146. Springer, 2020.

\bibitem{NIPS2012_c399862d}
Alex Krizhevsky, Ilya Sutskever, and Geoffrey~E Hinton.
\newblock Imagenet classification with deep convolutional neural networks.
\newblock In F. Pereira, C.~J.~C. Burges, L. Bottou, and K.~Q. Weinberger,
  editors, {\em Advances in Neural Information Processing Systems}, volume~25.
  Curran Associates, Inc., 2012.

\bibitem{8802509}
Mengchen Liu, Shixia Liu, Hang Su, Kelei Cao, and Jun Zhu.
\newblock Analyzing the noise robustness of deep neural networks.
\newblock In {\em 2018 IEEE Conference on Visual Analytics Science and
  Technology (VAST)}, pages 60--71, 2018.

\bibitem{lundberg2017unified}
Scott~M Lundberg and Su-In Lee.
\newblock A unified approach to interpreting model predictions.
\newblock {\em Advances in neural information processing systems}, 30, 2017.

\bibitem{metzen2017detecting}
Jan~Hendrik Metzen, Tim Genewein, Volker Fischer, and Bastian Bischoff.
\newblock On detecting adversarial perturbations.
\newblock {\em arXiv preprint arXiv:1702.04267}, 2017.

\bibitem{nguyen2020wide}
Thao Nguyen, Maithra Raghu, and Simon Kornblith.
\newblock Do wide and deep networks learn the same things? uncovering how
  neural network representations vary with width and depth.
\newblock {\em arXiv preprint arXiv:2010.15327}, 2020.

\bibitem{roth2019odds}
Kevin Roth, Yannic Kilcher, and Thomas Hofmann.
\newblock The odds are odd: A statistical test for detecting adversarial
  examples.
\newblock In {\em International Conference on Machine Learning}, pages
  5498--5507. PMLR, 2019.

\bibitem{selvaraju2017gradcam}
Ramprasaath~R Selvaraju, Abhishek Das, Ramakrishna Vedantam, Michael Cogswell,
  Devi Parikh, and Dhruv Batra.
\newblock Grad-cam: Why did you say that?, 2017.

\bibitem{simonyan2014very}
Karen Simonyan and Andrew Zisserman.
\newblock Very deep convolutional networks for large-scale image recognition.
\newblock {\em arXiv preprint arXiv:1409.1556}, 2014.

\bibitem{szegedy2016rethinking}
Christian Szegedy, Vincent Vanhoucke, Sergey Ioffe, Jon Shlens, and Zbigniew
  Wojna.
\newblock Rethinking the inception architecture for computer vision.
\newblock In {\em Proceedings of the IEEE conference on computer vision and
  pattern recognition}, pages 2818--2826, 2016.

\bibitem{szegedy2014intriguing}
Christian Szegedy, Wojciech Zaremba, Ilya Sutskever, Joan Bruna, Dumitru Erhan,
  Ian Goodfellow, and Rob Fergus.
\newblock Intriguing properties of neural networks, 2014.

\bibitem{vakhshiteh2020adversarial}
Fatemeh Vakhshiteh, Ahmad Nickabadi, and Raghavendra Ramachandra.
\newblock Adversarial attacks against face recognition: A comprehensive study.
\newblock {\em arXiv preprint arXiv:2007.11709}, 2020.

\bibitem{NEURIPS2021_3937230d}
Boxi Wu, Jinghui Chen, Deng Cai, Xiaofei He, and Quanquan Gu.
\newblock Do wider neural networks really help adversarial robustness?
\newblock In M. Ranzato, A. Beygelzimer, Y. Dauphin, P.S. Liang, and J.~Wortman
  Vaughan, editors, {\em Advances in Neural Information Processing Systems},
  volume~34, pages 7054--7067. Curran Associates, Inc., 2021.

\bibitem{Xie_2019_CVPR}
Cihang Xie, Yuxin Wu, Laurens van~der Maaten, Alan~L. Yuille, and Kaiming He.
\newblock Feature denoising for improving adversarial robustness.
\newblock In {\em Proceedings of the IEEE/CVF Conference on Computer Vision and
  Pattern Recognition (CVPR)}, June 2019.

\bibitem{9286885}
Chongzhi Zhang, Aishan Liu, Xianglong Liu, Yitao Xu, Hang Yu, Yuqing Ma, and
  Tianlin Li.
\newblock Interpreting and improving adversarial robustness of deep neural
  networks with neuron sensitivity.
\newblock {\em IEEE Transactions on Image Processing}, 30:1291--1304, 2021.

\end{thebibliography}

}

\end{document}